\documentclass{ecai} 

\usepackage{latexsym}
\usepackage{amssymb}
\usepackage{amsmath}
\usepackage{amsthm}
\usepackage{booktabs}
\usepackage{enumitem}
\usepackage{graphicx}
\usepackage{color}

\usepackage{colortbl}
\usepackage{enumitem}
\usepackage{makecell}
\usepackage{float}
\usepackage{xcolor}

\usepackage{multirow}
\usepackage{array}
\usepackage[implicit=false]{hyperref}

\definecolor{green}{HTML}{DAF2D0}
\definecolor{red}{HTML}{FFB7AF}
\definecolor{grey}{HTML}{D0D0D0}
\definecolor{darkred}{HTML}{FF0000}
\definecolor{darkgrey}{HTML}{71797E}
\definecolor{lightgrey}{HTML}{D3D3D3}

\newcommand{\BibTeX}{B\kern-.05em{\sc i\kern-.025em b}\kern-.08em\TeX}

\begin{document}


\begin{frontmatter}


\paperid{1173} 


\title{Score Before You Speak: Improving Persona Consistency in Dialogue Generation using Response Quality Scores}


\author[A]{\fnms{Arpita}~\snm{Saggar}\thanks{Corresponding Author. Email: scasag@leeds.ac.uk}}
\author[B]{\fnms{Jonathan C.}~\snm{Darling}}
\author[A]{\fnms{Vania}~\snm{Dimitrova}}
\author[A]{\fnms{Duygu}~\snm{Sarikaya}} 
\author[A]{\fnms{David C.}~\snm{Hogg}}

\address[A]{School of Computer Science, University of Leeds}
\address[B]{Leeds Institute of Medical Education, School of Medicine, University of Leeds}


\begin{abstract}
Persona-based dialogue generation is an important milestone towards building conversational artificial intelligence. Despite the ever-improving capabilities of large language models (LLMs), effectively integrating persona fidelity in conversations remains challenging due to the limited diversity in existing dialogue data. We propose a novel framework SBS (Score-Before-Speaking), which outperforms previous methods and yields improvements for both million and billion-parameter models. Unlike previous methods, SBS unifies the learning of responses and their relative quality into a single step. The key innovation is to train a dialogue model to correlate augmented responses with a quality score during training and then leverage this knowledge at inference. We use noun-based substitution for augmentation and semantic similarity-based scores as a proxy for response quality. Through extensive experiments with benchmark datasets (PERSONA-CHAT and ConvAI2), we show that score-conditioned training allows existing models to better capture a spectrum of persona-consistent dialogues. Our ablation studies also demonstrate that including scores in the input prompt during training is superior to conventional training setups. Code and further details are available at \href{https://arpita2512.github.io/score_before_you_speak}{https://arpita2512.github.io/score\_before\_you\_speak}.
\end{abstract}

\end{frontmatter}


\section{Introduction}

Building intelligent dialogue agents that can mimic human conversational abilities has been an important milestone in advancing artificial intelligence. This requires agents to maintain a consistent persona throughout the interaction, in order to enhance engagement and gain the trust of users \citep{Zheng_Zhang_Huang_Mao_2020}. A persona is captured in sentences describing an individual's personality and background information. Ensuring persona consistency is challenging since it requires responses to be relevant to the dialogue context as well as the persona. Various approaches have been proposed to incorporate personas into dialogue generation. Most existing frameworks, such as ORIG \citep{orig} and SimOAP \citep{zhou2023a}, use task-agnostic pretrained language models that require large amounts of persona-specific data for finetuning. However, persona-based dialogues have limited availability and diversity due to the costs associated with curating them. These datasets often rely on pairs of crowd-sourced workers to play the role of the assigned persona while conversing with each other \citep{zhang2018}, which is expensive and time-consuming to collect. Compared to conventional conversational datasets like DailyDialog \citep{li-etal-2017-dailydialog}, persona-based dialogues are relatively scarce and less diverse. This has precipitated the creation of innovative training frameworks that can maximise the amount of information learnt from persona-based conversations. Many existing approaches to train dialogue generation models traditionally use a two-stage pipeline, which involves supervised finetuning followed by aligning augmented responses with preference signals. Within the subdomain of persona-consistent dialogue generation, alignment with perceived quality (judged by a critic model or using metrics) is achieved through methods like contrastive learning \citep{li2023b}, reinforcement learning \citep{liu2020, Song_Zhang_Hu_Liu_2020, shea-yu-2023-building}, or self-guided retrieval augmented generation \citep{huang-etal-2023-learning}. While this standardised pipeline produces impressive results, accurately assessing the relative quality of outputs is non-trivial. The computational complexity of these approaches also reduces their applicability in resource-constrained environments. 

We present a new framework SBS (Score-Before-Speaking), which unifies learning desirable dialogue responses and their relative quality into a single step while outperforming previous work. We first address the issue of limited diversity by using linguistic knowledge to mask and regenerate nouns in dialogues, which creates an augmented dataset. This approach is motivated by the idea that persona descriptions are largely characterised by nouns \citep{fraurud1992}. Next, we score the augmented responses based on semantic closeness to the original, gold-standard responses. These scores act as a proxy for response quality and are incorporated into the input prompt while training, to teach the model a mapping between responses and quality scores. We show that augmented responses can have varying levels of semantic closeness to the original data, which motivates the idea of utilising them for quality alignment. Our method only needs a single learning objective to train a persona-consistent dialogue model. Experimental results with the PERSONA-CHAT and ConvAI2 datasets illustrate that our framework improves current baselines on both automatic and human metrics. Our contributions are summarised below: 
\begin{enumerate}
    \item A framework for improving persona consistency in dialogue through training conditioned on quality scores.
    \item A comparative evaluation demonstrating state-of-the-art performance with both small and large language models.
    \item An ablation study highlighting the effectiveness of incorporating quality scores in the input prompt during training.
\end{enumerate}

\begin{figure*}
  \includegraphics[width=\linewidth]{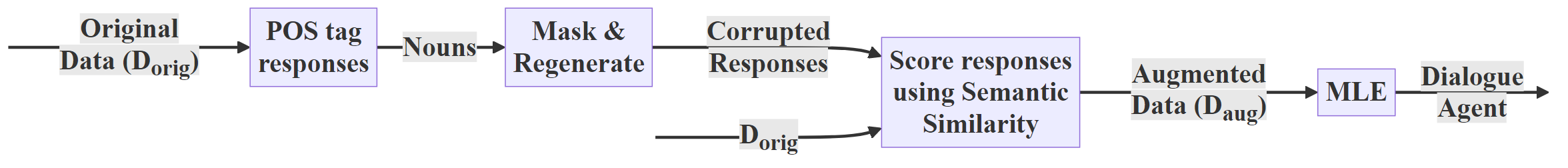}
  \caption{An overview of the SBS framework. Part-of-speech tagging is performed for responses in the original (gold-standard) dataset. The nouns in responses are then masked and regenerated (one at a time) to create corrupted responses. Corrupted responses are scored based on their semantic similarity to the original responses, while the original responses all receive a score of 1.0. Finally, a decoder model is trained via MLE to produce the dialogue model.}
  \label{fig:overview}
\end{figure*}


\section{Related Work}

\subsection{Data Augmentation}

Owing to their limited diversity, persona-based dialogues are more difficult to learn as compared to traditional dialogues. Responses can have varying levels of conformity to the assigned persona \citep{wakaki2024}, which is often not captured fully in the data and makes it difficult to reliably map responses to personas. These issues motivate the use of data augmentation techniques for training a robust persona-consistent dialogue agent \citep{cao2022}. Popular methods for creating lexical and syntactic paraphrases include back-translation \citep{sennrich-etal-2016-improving} and delexicalisation (replacing words with their semantic frames) followed by slot filling \citep{hou-etal-2018-sequence}. Others propose the use of external datasets for retrieval augmented generation \citep{huang-etal-2023-learning}. Many recent works leverage large language models (LLMs) to generate synthetic data \citep{shao-etal-2023-character, wang-etal-2024-rolellm, ran-etal-2024-capturing}. 

Approaches for generating negative samples include randomly selecting responses from different conversations \citep{zhang2018}, applying syntactic transformations to swap the object and subject \citep{min-etal-2020-syntactic}, and token or sentence-level manipulations \citep{zhou2021, steindl-etal-2024-counterfactual}. Our work differs from previous methods by targeting nouns for data manipulation and using a regression-based score to quantify how close the augmented samples are to the original dataset.

\subsection{Persona-based Dialogue Generation}

Much work has been done to personalise open-domain dialogue generation. TransferTransfo \citep{wolf_2019} combines pretraining with multi-task finetuning to achieve state-of-the-art results in the ConvAI2 competition \citep{dinan2019}, a challenge aimed at creating dialogue agents capable of articulating open-domain conversation. Others use reinforcement learning with pseudo-negative examples created by swapping dialogues and personas \citep{takayama-etal-2025-persona} or additional loss terms for predicting keywords in responses \citep{bok}. While earlier work utilises simpler models like RNNs or LSTMs, most current methods make use of Transformer-based \citep{vaswani_2017} GPT-style architectures \citep{Radford2019LanguageMA}, such as the Llama models by Meta \citep{llama1, llama2, llama3}. More sophisticated, multi-stage approaches perform supervised fine-tuning through a curriculum learning approach \citep{cao2022}, or in combination with unlikelihood training \citep{kim2023}. CLV \citep{tang-etal-2023-enhancing-personalized} uses a contrastive learning approach to learn latent representations of persona description pieces. These are then fed into a decider network to select the most preferred persona pieces during generation. The DITTO framework \citep{lu-etal-2024-large} reformulates role-playing as a reading comprehension task and generates a dataset using profiles from open-access datasets. This self-generated data is then used to train models. Another method is Selective Prompt Tuning \citep{huang-etal-2024-selective}, which first trains a retriever network to rank multiple soft prompts for a given context. Contrastive learning with incorrect contexts is then used to update the soft prompts, and the highest-ranked prompts are finally used to generate dialogues. In contrast, our work directly incorporates quality scores in the input prompt while training, eliminating the need to finetune any intermediate models.


\section{Methodology}
\label{sec:method}

In this work, we propose a single objective training framework (SBS) for generating persona-consistent dialogues. As shown in Figure \ref{fig:overview}\footnote{Figures \ref{fig:overview}, \ref{fig:mask} and \ref{fig:inf} have been created using Mermaid \cite{Sveidqvist_Mermaid_Generate_diagrams_2014}.}, we begin with data augmentation, where nouns in dialogue responses are masked and then regenerated to create corrupted samples. Next, we score the corrupted responses by quantifying their level of semantic closeness to the original responses. All original responses are assigned the maximum score. Prior work advocates the use of thresholding to filter augmented data \citep{zhang-etal-2020-dialogue, wu-etal-2022-dg2}. However, using a threshold value can obscure the distinction between samples with scores near the threshold and those with scores significantly away from it. Therefore, we do not perform any filtering for the augmented data and instead incorporate scores into the input sequence during training. By conditioning response generation on scores, we teach the model to learn a fine-grained mapping between responses and their quality (the level of corruption with respect to the original response). Responses with high scores help the model learn desirable dialogues, while those with low scores (persona-inconsistent) help delineate areas of inconsistency through regeneration-driven contradiction and irrelevance. Finally, we train the dialogue model using our augmented dataset, where each sample contains a persona, a dialogue history, a user query\footnote{The last utterance by the user, which elicits a response but is not necessarily a question.}, a response and a score for the response.

\subsection{Data Augmentation}

\begin{figure}
  \includegraphics[width=0.99\columnwidth]{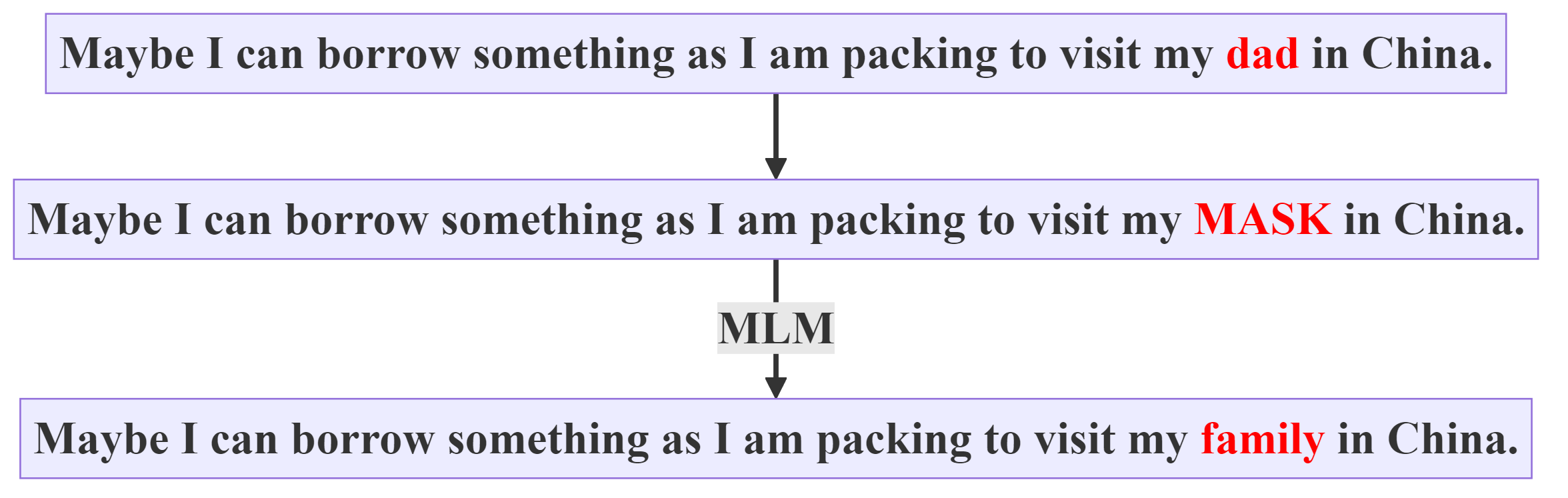}
  \caption{An example from the PERSONA-CHAT train set where the regenerated response is a paraphrase of the original response and is not inconsistent with the relevant persona sentence: \emph{My father lives in China.}}
  \label{fig:mask}
\end{figure}

We first formally define our persona-based dialogue dataset $D_{orig}$. For each $D_i = (P_i, H_i, Q_i, R_i)$ in $D_{orig}$, where $1\le i \le|D_{orig}|$, we have $n_i$ sentences that capture the speaker's persona $P_i = \{p_i^1, .  . , p_i^{n_i}\}$, a history with $m_i$ previous utterances $H_i = \{h_i^1, . . , h_i^{m_i}\}$, a user query $Q_i$, and a golden, persona-consistent response $R_i$. Note that $n_i$ and $m_i$ need not be the same for all samples.

To capture a fine-grained spectrum of persona-consistency, we want augmented samples to retain some relevance to the dialogue and have similar style and tone. Therefore, we choose masked language modelling for data augmentation, which provides greater controllability compared to LLM-based augmentation. For mask selection, previous approaches use sampling \citep{zhou2021}, entailment scores \citep{li2023b}, and differences in word probabilities estimated by a masked language model (MLM) with and without the dialogue history \citep{park2021}. We simplify this selection using linguistic concepts. We note that persona descriptions largely consist of sentences containing referential attributes such as occupation, likes and dislikes, and so on. These attributes manifest in discourse as referential expressions \citep{vonk1992}, the most important of which is generally the noun phrase \citep{fraurud1992, Khatri2018}. Furthermore, nouns are the most common type of lexical word in sentences \citep{stoian2022language}, and their primary purpose is to profile things \citep{Hollmann+2013+275+308}. Using this information, we hypothesise that nouns capture the most persona-relevant information and we choose to mask the nouns present in responses. We perform part-of-speech (POS) tagging and regenerate masked nouns using an MLM, without providing any context (persona and dialogue history). Nouns are masked one at a time to ensure that some relevance to the original data persists. If the MLM predicts the original word that was masked, then the second most likely word is selected.

While prior work treats regenerated responses as negative, i.e., persona-inconsistent \citep{park2021, li2023b}, our analysis reveals that this is not always true. While some regenerated responses do contradict the assigned persona, many are paraphrases of the original response (Figure \ref{fig:mask}). To account for this variability, we use BERTScore \citep{bert-score} to score regenerated responses relative to the original responses. It works by computing token-level cosine similarities between contextual embeddings. We use the DeBERTa-XLarge-MNLI model \citep{he2021deberta} to compute scores since it has the best correlation with human evaluation\footnote{https://github.com/Tiiiger/bert\_score}. The golden responses from $D_{orig}$ are all assigned the value $1.0$\footnote{The highest possible value for cosine similarity}. We add these new examples along with the scores for all responses to $D_{orig}$ to create our augmented dataset, $D_{aug}$, where each $D_i$ contains an additional attribute $S_i$ that corresponds to the score for that sample's response $R_i$. 

\subsection{Training} 

We train our dialogue agent via Maximum Likelihood Estimation (MLE) and use our augmented dataset $D_{aug}$. Given a training sample ($P_i, H_i, Q_i, S_i, R_i$), we concatenate $P_i, H_i, Q_i$ and $S_i$ (up to 2 decimal places) to form the input sequence that is used to predict the target response $R_i$. The training loss is then computed as:
\begin{equation}
    \label{eq:loss}
    L_{MLE} = \frac{-1}{T_i}\sum\limits_{j=1}^{T_i} \log p_{\theta}(R_i^j|P_i, H_i, Q_i, S_i, R_i^{<j})
\end{equation}
where $T_i$, which is indexed by $j$, is the number of tokens in $R_i$, $R_i^{<j}$ represents all tokens preceding the $j^{th}$ token in $R_i$, and $\theta$ represents the parameters of the dialogue agent. The model learns to predict responses conditioned on their respective personas, dialogue histories, user queries and scores. Our goal is to teach the dialogue agent a correspondence between score and response quality so that this knowledge can then be exploited during inference. An analysis of the correspondence learnt is presented in Section \ref{sec:scoring}.

\section{Experiments}

To validate the effectiveness of our approach, we conduct experiments with two popular English language datasets, PERSONA-CHAT \citep{zhang2018} and ConvAI2 \citep{dinan2019}, sourced from the ParlAI framework \citep{miller2017parlai}. PERSONA-CHAT consists of conversations between randomly paired crowd workers portraying specific personas. The dataset provides 1155 personas for training, each with at least 4 sentences, setting aside 100 separate personas for validation, and 100 for testing. PERSONA-CHAT contains 65,719 query-response pairs for training (189,294 post augmentation), 7801 for validation and 7512 for testing. ConvAI2 extends PERSONA-CHAT by crowd-sourcing related sentences (rephrases, generalisations and specialisations) to make the task more challenging. It has 131,438 dialogue pairs for training (373,545 post augmentation) and 7801 for validation. Since the test set of ConvAI2 is not publicly available, we test using its validation set and validate using a subset (10\%) of its training set. The Stanza toolkit \citep{qi2020} is used to perform POS tagging and the BART encoder-decoder model \citep{lewis2020} to regenerate masked nouns. For finetuning, we select two popular language models that have been optimised for conversational response generation - DialoGPT (117M parameters) \citep{zhang2020a} and Llama 3.1 Instruct (8B parameters) \citep{llama3}. We use Hugging Face's Transformers library \citep{wolf2020} to implement our framework and the Optuna library \citep{akiba2019optuna} for hyperparameter tuning. The Llama models are trained using Low-Rank Adaptation (LoRA) \citep{hu2022lora}. Further training details are provided in the supplementary material \citep{supplementary}.

We employ standard natural language generation metrics for evaluation. For faithfulness to the original responses, we use Perplexity \citep{zhang2018} (the exponentiated average negative log-likelihood of a sequence) and BLEU \citep{papineni2002} (n-gram overlap; SacreBLEU \citep{post-2018-call} implementation). To measure the diversity of generated responses, we use Distinct-n \citep{li2016} (the ratio of distinct n-grams to the total n-grams) and Entropy-n \citep{zhang2018a} (modification of Distinct-n which incorporates n-gram distribution in its calculation). We also use Consistency Score (C) \citep{madotto2019}, which uses natural language inference to indicate the consistency between generated responses and the assigned persona. For all the metrics mentioned above, a higher value indicates better performance, except for perplexity, where lower values are desirable.

\begin{figure}
  \includegraphics[width=0.95\columnwidth]{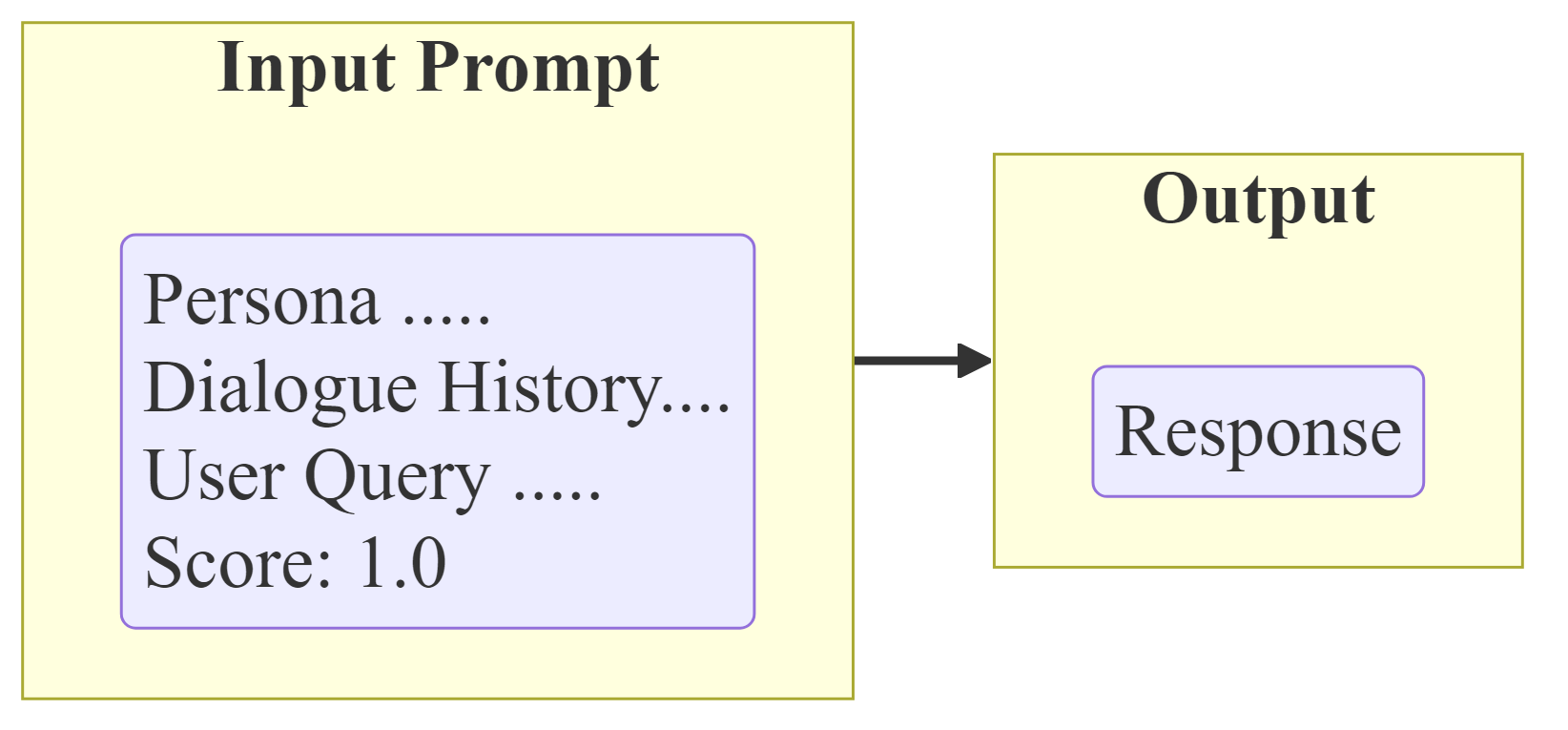}
  \caption{Input and output showing how scores are exploited during inference. We train the model to correlate scores (range [-1, 1]) with responses. This mapping is then leveraged at inference by including a score of 1.0 in the input prompt to generate high-quality responses.}
  \label{fig:inf}
\end{figure}

\textbf{Inference} Since our models are trained to generate responses conditioned on scores, we leverage this knowledge at inference. Response generation is done by explicitly setting a score of 1.0 (the highest score in the train set) in the input prompt, as shown in Figure \ref{fig:inf}. Beam search with beam size 3 is used to generate responses.

\subsection{Compared Methods} 

We compare experimental results produced by our framework with the DialoGPT and Llama 3.1 baselines (models finetuned with the original dataset), and other state-of-the-art methods:
\begin{itemize}
    \item \textbf{Learning to Know Myself (LTKM)} \citep{li2023b}: A two-stage framework that first trains a dialogue agent with persona-relevant responses, followed by contrastive learning with persona-inconsistent responses.
    \item \textbf{GPT2-D3} \citep{cao2022}: A curriculum learning approach which first trains the dialogue agent with an easier, distilled version of the data, followed by the original data.
    \item \textbf{Persona-Adaptive Attention (PAA)} \citep{huang2023}: A method that models the persona and dialogue context using two separate encoder models, then fuses these with a decoder through dynamic masking.
    \item \textbf{ORder Insensitive Generation (ORIG)} \citep{orig}: An approach which optimises models under the constraint that response generation should be invariant to different orderings of persona sentences.
    \item \textbf{BoK} \citep{bok}: A framework which introduces a novel Bag-of-Keywords (BoK) loss to capture the core components of the next utterance using keyword prediction. 
    \item \textbf{Selective Prompt Tuning (SPT)} \citep{huang-etal-2024-selective}: A modification of prompt tuning \citep{lester-etal-2021-power} that trains a dense retriever to adaptively select the best soft prompt from a group based on the input.
    \item \textbf{Learning Retrieval Augmentation for Personalized DialOgue Generation (LAPDOG)} \citep{huang-etal-2023-learning}: A two-stage method that leverages external story data for persona dialogue generation. A generator is trained in the first stage and then jointly optimised with a retriever in the second stage.
\end{itemize}

\begin{table*}[h]
    \centering
    \begin{tabular}{lllccccccccccc}
        \specialrule{1.6pt}{1pt}{1pt}
        
        & \textbf{Method} & \textbf{Model} & \textbf{Size} & \textbf{PPL} & \textbf{BLEU-1} & \textbf{BLEU-2} & \textbf{BLEU-3} & \textbf{BLEU-4} & \textbf{Dist-1} & \textbf{Dist-2} & \textbf{Ent-1} & \textbf{Ent-2} & \textbf{C} \\ 
        \cline{2-14}        
        \multirow{9}*{(a)} & Baseline & DialoGPT & 117M &\underline{13.12} &20.68 &10.49 &5.73 &3.31 &3.78 &12.51 &4.16 &6.59 &56.23 \\
        \cline{2-14}
        & D3 \citeyearpar{cao2022} & GPT2 &117M &15.69 &- &- &- &4.18 &2.27 &9.80 &4.21 &6.43 &55.70 \\
        \cline{2-14}
        & \multirow{4}*{LTKM \citeyearpar{li2023b}} &Seq2Seq &- &29.87 &21.70 &11.41 &6.51 &4.03 &1.67 &6.53 &4.27 &5.81 &41.72 \\
        & &Transformer &213M &33.75 &19.82 &9.85 &5.37 &3.18 &1.33 &5.64 &4.15 &5.83 &28.71 \\
        & &GPT2 &117M &14.87 &19.91 &10.35 &6.13 &3.86 &2.35 &9.73 &\textbf{5.04} &6.54 &62.22 \\
        & &DialoGPT &117M &13.44 &20.59 &10.67 &6.28 &3.95 &3.39 &10.85 &4.95 &6.40 &61.97 \\
        \cline{2-14}
        & ORIG \citeyearpar{orig} &GPT2 &117M &- &\underline{23.14} &\underline{12.42} &7.16 &4.34 &\textbf{4.33} &\textbf{14.60} &4.97 &\textbf{6.79} &\underline{62.49} \\
        \cline{2-14}
        & BoK \citeyearpar{bok} &DialoGPT &117M &14.31 &21.58 &12.36 &\underline{7.44} &\underline{4.59} &3.42 &11.46 &4.87 &6.43 &44.91 \\
        \cline{2-14}
        & \textbf{SBS (Ours)} &DialoGPT &117M &\textbf{11.92} &\textbf{23.65} &\textbf{12.81} &\textbf{7.53} &\textbf{4.71} &\underline{4.22} &\underline{13.96} &\underline{4.99} &\underline{6.63} & \textbf{66.64} \\
        
        \specialrule{1.6pt}{1pt}{1pt}

        \multirow{8}*{(b)} & Baseline & DialoGPT &117M &\underline{13.30} &17.99 &8.62 &4.91 &2.84 &3.65 &15.67 &5.06 &6.56 &60.84 \\
        \cline{2-14}
        & \multirow{4}*{LTKM \citeyearpar{li2023b}} &Seq2Seq &- &27.52 &20.54 &\underline{10.46} &\underline{6.38} &\underline{3.92} &2.01 &8.20 &4.32 &5.96 &46.88 \\
        & &Transformer &213M &30.63 &19.33 &9.69 &5.79 &3.51 &1.36 &5.76 &4.21 &5.61 &39.84 \\
        & &GPT2 &117M &14.41 &18.95 &9.20 &5.43 &3.36 &2.35 &10.26 &5.22 &6.77 &54.56 \\
        & &DialoGPT &117M &\textbf{13.19} &19.70 &9.85 &5.70 &3.52 &2.37 &11.14 &\textbf{5.39} &\underline{7.10} &63.90 \\
        \cline{2-14}
        & PAA \citeyearpar{huang2023} &GPT2 &254M &14.03 &\underline{21.20} &10.10 &5.52 &3.20 &2.89 &9.25 &4.96 &6.70 &\underline{68.16} \\
        \cline{2-14}
        & SPT \citeyearpar{huang-etal-2024-selective} & OPT &125M &- &20.63 &9.86 &5.56 &3.22 &\textbf{4.87} &\textbf{17.38} &5.28 &\textbf{7.29} &65.13 \\
        \cline{2-14}
        & \textbf{SBS (Ours)} &DialoGPT & 117M&14.02 &\textbf{22.11} &\textbf{11.30} &\textbf{6.75} &\textbf{4.21} &\underline{4.69} &\underline{16.52} &\underline{5.29} &\underline{7.10} & \textbf{70.60} \\

        \specialrule{1.6pt}{1pt}{1pt}
    \end{tabular}
    \caption{Results comparing different frameworks applied to small language models for (a) PERSONA-CHAT test set, and (b) ConvAI2 validation set. Dist-n and C are in \%. The best results for each metric are in bold, while the second best are underlined.}
    \label{tab:results_small}
\end{table*}

\begin{table*}[h]
    \centering
    \begin{tabular}{lllccccccccccc}
        \specialrule{1.6pt}{1pt}{1pt}
        
        & \textbf{Method} & \textbf{Model} & \textbf{Size} & \textbf{PPL} & \textbf{BLEU-1} & \textbf{BLEU-2} & \textbf{BLEU-3} & \textbf{BLEU-4} & \textbf{Dist-1} & \textbf{Dist-2} & \textbf{Ent-1} & \textbf{Ent-2} & \textbf{C} \\ 
        \cline{2-14}        
        \multirow{3}*{(a)} & Baseline & Llama 3.1 &8B &\textbf{6.71} &23.88 &12.46 &7.05 &4.20 &5.15 &18.91 &5.34 &7.34 &57.93 \\
        \cline{2-14}
        & \textbf{SBS (Ours)} &Llama 3.1 &8B &6.88 &\textbf{26.27} &\textbf{14.72} &\textbf{9.00} &\textbf{5.85} &\textbf{5.18} &\textbf{19.50} &\textbf{5.37} &\textbf{7.49} &\textbf{64.47} \\
        
        \specialrule{1.6pt}{1pt}{1pt}

        \multirow{4}*{(b)} & Baseline & Llama 3.1 &8B &\textbf{6.63} &21.57 &10.94 &6.43 &3.88 &\underline{5.81} &22.43 &\textbf{5.73} &7.87 &64.07 \\
        \cline{2-14}
        & SPT \citeyearpar{huang-etal-2024-selective} & Llama 2 &7B &- &17.67 &9.02 &5.27 &3.21 &5.29 &22.08 &\underline{5.69} &\textbf{8.30} &\underline{69.77} \\
        \cline{2-14}
        & LAPDOG \citeyearpar{huang-etal-2023-learning} & T5 & 11B &- &\underline{22.48} &\underline{11.41} &\underline{6.73} &\underline{4.14} &\textbf{5.88} &\textbf{24.76} &5.27 &\underline{8.15} &59.92 \\
        \cline{2-14}
        & \textbf{SBS (Ours)} &Llama 3.1 &8B &6.98 &\textbf{25.22} &\textbf{13.12} &\textbf{7.82} &\textbf{4.99} &5.71 &\underline{23.09} &5.68 &7.96 &\textbf{73.11}  \\

        \specialrule{1.6pt}{1pt}{1pt}
    \end{tabular}
    \caption{Results comparing different frameworks applied to large language models for (a) PERSONA-CHAT test set, and (b) ConvAI2 validation set. Dist-n and C are in \%. The best results for each metric are in bold, while the second best are underlined.}
    \label{tab:results_llm}
\end{table*}

\subsection{Results}
\label{subsec:results}

We report the performance of our models using the test set of PERSONA-CHAT and the validation set of ConvAI2. Table \ref{tab:results_small} presents metrics for small models (< 1 billion parameters), and Table \ref{tab:results_llm} contains the results for large language models. For PERSONA-CHAT, our method improves the Llama 3.1 baseline on nearly all metrics and surpasses or matches previous state-of-the-art works for smaller models. For ConvAI2, we again achieve similar or better performance for both small and large models, barring the diversity metrics (Dist-n and Ent-n) for LLMs. Another point to consider is the training effort required to implement various frameworks. LTKM, BoK and ORIG optimise a second objective (Contrastive loss, Bag-of-Keywords loss and KL Divergence respectively) in addition to MLE to train their models. GPT2-D3 finetunes five extra models (one for data distillation, two for data diversification, one for response alignment and one for quality filtering) as part of its pipeline. PAA trains two separate encoder models to capture persona and context. Both the LAPDOG AND SPT frameworks train an additional model for retrieval. In contrast, our approach only needs a single model and a single training objective, though the supervised training effort likely increases due to the presence of additional score tokens.

\textbf{Human Evaluation}  We randomly select 100 samples from the datasets (50 each) to compare our framework with the baseline models (finetuned with the original, non-augmented datasets). We recruited four evaluators to assess the generated responses. These are native speakers who are proficient in the English language but do not have any details about the models or training processes. Each evaluator rated responses on three aspects: coherence (relevance to context), persona-consistency (relevance to persona) and fluency (correct syntax and grammar). Each of these aspects is measured on an ordinal scale ranging from 1 (lowest) to 3 (highest). These are:
\begin{itemize}
    \item \textbf{Coherence:} 1 (irrelevant to dialogue context), 2 (somewhat relevant to dialogue context), 3 (relevant to dialogue context)
    \item \textbf{Persona Consistency:} 1 (Directly or indirectly contradicts persona), 2 (Neither reflects nor contradicts persona), 3 (Directly or indirectly reflects persona)
    \item \textbf{Fluency:} 1 (Broken text, difficult to follow), 2 (Some errors but understandable), 3 (Grammatically correct, easy to understand)
\end{itemize}

The results for the human evaluation are (baseline vs SBS; Wilcoxon signed-rank test p-value): coherence (2.35 vs 2.61; $p=1e-3$), persona-consistency (2.58 vs 2.86; $p=5.5e-5$) and fluency (2.90 vs 2.98; $p=1.1e-2$).  SBS significantly improves on the baseline in all three aspects, which is consistent with the results of the automatic evaluation. The agreement rates for the evaluators are 67\% (coherence), 87.5\% (persona consistency) and 92\% (fluency) @3, i.e., at least three of them reach an agreement. These results indicate the validity of our evaluation while also highlighting the subjectivity inherent in assessing natural language. 

\subsection{Ablation Study}
\label{sec:ablation}

To evaluate how the presence of scores impacts performance, we conduct ablation tests by finetuning models without adding scores in the input sequence. We train using the augmented data under two settings: (1) no score in the input prompt and no thresholding based on scores, and (2) no score in the input prompt with data thresholded using similarity scores. A limited set of scores is chosen for the thresholded setting since the majority (75\%) of scores for corrupted (masked and regenerated) responses lie between $0.70$ and $1.0$ for both datasets. Table \ref{tab:res_ablation} present the results of the ablation tests for Llama 3.1 (the results for DialoGPT are similar and are presented in the supplementary material \citep{supplementary}). For both datasets, SBS (scores in the input prompt and no thresholding) produces either the best or second best results across nearly all metrics, highlighting the benefit of including scores in the input while training. Notably, while trends for most metrics vary, there is a (almost) steady degradation in consistency score (C) as more corrupted samples are added to the training set without any corresponding information to indicate quality.

\begin{table*}[h]
    \small
    \centering
    \begin{tabular}{lccccccccccc}
        \specialrule{1.6pt}{1pt}{1pt}
        
        & \textbf{Method} & \textbf{PPL} & \textbf{BLEU-1} & \textbf{BLEU-2} & \textbf{BLEU-3} & \textbf{BLEU-4} & \textbf{Dist-1} & \textbf{Dist-2} & \textbf{Ent-1} & \textbf{Ent-2} &\textbf{C} \\ 
        \cline{2-12}
        
        \multirow{7}*{(a)} & \makecell{Score in Prompt +\\No Threshold (Ours)} &\textbf{6.88} &\textbf{26.27} &\textbf{14.72} &\textbf{9.00} &\textbf{5.85} &5.18 &\underline{19.50}	&5.37	&\underline{7.49}	&\textbf{64.47}\\
        \cline{2-12}
        & \makecell{No Score in Prompt\\+ Threshold=0.95} &8.16	&25.38	&14.07	&8.47	&5.34	&5.12	&18.16	&5.37	&7.37 &\underline{52.98}\\
        \cline{2-12}
        & \makecell{No Score in Prompt\\+ Threshold=0.90} &8.04	&\underline{26.08}	&14.52	&8.38	&5.26	&5.12	&18.93	&\underline{5.41}	&\underline{7.49} &48.20\\
        \cline{2-12}
        & \makecell{No Score in Prompt\\+ Threshold=0.85} &8.01	&26.04	&\underline{14.67}	&\underline{8.95}	&\underline{5.78}	&5.07	&18.36	&5.33	&7.34 &46.80\\
        \cline{2-12}
        & \makecell{No Score in Prompt\\+ Threshold=0.80} &7.98	&25.67	&14.32	&8.67	&5.55	&5.11	&18.35	&5.34	&7.32 &44.86\\
        \cline{2-12}
        & \makecell{No Score in Prompt\\+ Threshold=0.75} &\underline{7.93}	&22.23	&12.81	&8.00	&4.99	&\textbf{5.78}	&19.42	&\textbf{5.44} &7.36 &41.50\\
        \cline{2-12}
        & \makecell{No Score in Prompt\\+ No Threshold} &7.99	&22.13	&12.14	&7.33	&4.69	&\underline{5.56}	&\textbf{20.99}	&5.40	&\textbf{7.52} &43.16\\
        
        \specialrule{1.6pt}{1pt}{1pt}

        \multirow{7}*{(b)} & \makecell{Score in Prompt +\\No Threshold (Ours)} &6.98	&\textbf{25.22}	&\textbf{13.12}	&\textbf{7.82}	&\underline{4.99}	&\underline{5.71}	&\textbf{23.09}	&\underline{5.68}	&\textbf{7.96}	&\textbf{73.11}\\
        \cline{2-12}
        & \makecell{No Score in Prompt\\+ Threshold=0.95} &\textbf{6.67}	&22.08	&11.60	&7.04	&4.58	&\textbf{5.77}	&21.52	&\textbf{5.70}	&7.84 &\underline{57.03}\\
        \cline{2-12}
        & \makecell{No Score in Prompt\\+ Threshold=0.90} &6.76	&23.96	&13.00	&\textbf{7.82}	&4.09	&5.51	&20.73	&5.63	&7.78 &55.01\\
        \cline{2-12}
        & \makecell{No Score in Prompt\\+ Threshold=0.85} &\underline{6.74}	&22.35	&\underline{13.10}	&7.03	&4.54	&5.68	&\underline{22.33}	&5.64	&7.90 & 56.80\\
        \cline{2-12}
        & \makecell{No Score in Prompt\\+ Threshold=0.80} &6.77	&\underline{24.33}	&12.88	&\underline{7.77}	&\textbf{5.03}	&5.34	&20.92	&5.66	&\underline{7.92} &53.74\\
        \cline{2-12}
        & \makecell{No Score in Prompt\\+ Threshold=0.75} &6.80	&23.11	&12.54	&7.28	&4.54	&5.51	&20.58	&5.63	&7.88 &52.28\\
        \cline{2-12}
        & \makecell{No Score in Prompt\\+ No Threshold} &6.81	&22.98	&12.17	&7.39	&4.82	&5.56	&20.99	&5.59	&7.74 & 51.13\\

        \specialrule{1.6pt}{1pt}{1pt}
    \end{tabular}
    \caption{Ablation tests showing the effect of removing score tokens from the input during training for (a) PERSONA-CHAT test set, and (b) ConvAI2 validation set. Dist-n and C are in \%. The best results for each metric are in bold, while the second best are underlined.}
    \label{tab:res_ablation}
\end{table*}

\subsection{Influence of Score on Generation}
\label{sec:scoring}

Since our framework uses semantic similarity as a proxy for the response quality, we expect that the dialogue model will learn to correlate responses with scores during training. To investigate how well the models have learnt this relationship, we generate responses using lower scores in the input prompt. The expected trend is that lower scores should lead to lower-quality responses and therefore poorer metrics. Similar to the ablation study, we only use scores which compose the majority of augmented samples, since it is unlikely that the model learns any meaningful correspondence for smaller values. Table \ref{tab:scoring} reports the performance of metrics for DialoGPT using different scores in the input prompt for both datasets. The trends for Llama 3.1 are very similar and are presented in the supplementary material \citep{supplementary}. From Table \ref{tab:scoring}, we make three key observations:
\begin{itemize}
    \item The metrics follow the expected trend in most cases, as highlighted by the green cells, indicating that a relationship between score and response quality is learnt during training. 
    \item The biggest degradation in performance occurs when the score changes from 1.0 to 0.95, which is likely due to the nature of the training data. We note that the original training samples, which have a score of 1.0, make up nearly a third of the final augmented train set for both datasets. In contrast, the corrupted samples have scores distributed within the range of cosine similarity, rather than being concentrated around a specific value like $0.95$ or $0.90$.
    \item While it is unlikely that the dialogue agent will perfectly model the relationship between score and response quality (due to the nature of MLE), we anticipate that metrics should follow the expected trend most of the time. However, a contradiction is observed in Ent-1 and Ent-2 for PERSONA-CHAT. While greater consistency often leads to an acceptable diversity trade-off, a poorly trained model could be prone to neural text degeneration \citep{li2023repetition}. We investigate this result in section \ref{subsec:further} by analysing n-gram distributions. We also compute the diversity of responses in both datasets to examine why this contradiction occurs in only one case. 
\end{itemize}

\begin{table*}[h]
    \small
    \centering
    \begin{tabular}{cccccccccccc}
        \specialrule{1.6pt}{1pt}{1pt}
        
        & \textbf{Score} & \textbf{PPL} & \textbf{BLEU-1} & \textbf{BLEU-2} & \textbf{BLEU-3} & \textbf{BLEU-4} & \textbf{Dist-1} & \textbf{Dist-2} & \textbf{Ent-1} & \textbf{Ent-2} &\textbf{C} \\ 
        \cline{2-12}
        
        \multirow{6}*{(a)} & 1.0 &11.92	&23.65 &12.81 &7.53	&4.71 &4.22	&13.96 &4.99 &6.63 &66.64\\
        \cline{2-12}
        & 0.95 &\cellcolor{green}13.35 &\cellcolor{green}23.25 &{\cellcolor{green}}12.02 &\cellcolor{green}6.71 &\cellcolor{green}4.00 &\cellcolor{green}3.28 &\cellcolor{green}11.56 &\cellcolor{red}5.12 &\cellcolor{red}6.91 &\cellcolor{green}54.16 \\
        \cline{2-12}
        & 0.90 &\cellcolor{green}13.36 &\cellcolor{green}22.90 &\cellcolor{green}11.83 &\cellcolor{green}6.62 &\cellcolor{green}3.96 &\cellcolor{green}3.14 &\cellcolor{green}10.99 &\cellcolor{red}5.09 &\cellcolor{red}6.87 &\cellcolor{green}51.28 \\
        \cline{2-12}
        & 0.85 &\cellcolor{green}13.39 &\cellcolor{green}22.63 &\cellcolor{green}11.65 &\cellcolor{green}6.47 &\cellcolor{green}3.84 &\cellcolor{green}3.02 &\cellcolor{green}10.55 &\cellcolor{red}5.07 &\cellcolor{red}6.85 &\cellcolor{red}52.45 \\
        \cline{2-12}
        & 0.80 &\cellcolor{green}13.43 &\cellcolor{green}22.42 &\cellcolor{green}11.49 &\cellcolor{green}6.38 &\cellcolor{green}3.77 &\cellcolor{green}2.91 &\cellcolor{green}10.08 &\cellcolor{red}5.02 &\cellcolor{red}6.76 &\cellcolor{green}51.23 \\
        \cline{2-12}
        & 0.75 &\cellcolor{green}13.45 &\cellcolor{red}22.44 &\cellcolor{green}11.46 &\cellcolor{green}6.34 &\cellcolor{green}3.73 &\cellcolor{green}2.82 &\cellcolor{green}9.75 &\cellcolor{green}4.98 &\cellcolor{red}6.71 &\cellcolor{green}51.09 \\
        
        \specialrule{1.6pt}{1pt}{1pt}

        \multirow{6}*{(b)} & 1.0 &14.02 &22.11 &11.30 &6.75 &4.21 &4.69 &16.52 &5.29 &7.10 &70.60 \\
        \cline{2-12}
        & 0.95 &\cellcolor{green}15.62 &\cellcolor{red}22.27 &\cellcolor{green}10.79 &\cellcolor{green}6.06 &\cellcolor{green}3.58 &\cellcolor{green}3.60 &\cellcolor{green}13.34 &\cellcolor{green}5.28 &\cellcolor{red}7.17 &\cellcolor{green}56.98 \\
        \cline{2-12}
        & 0.90 &\cellcolor{green}15.67 &\cellcolor{green}21.87 &\cellcolor{green}10.49 &\cellcolor{green}5.87 &\cellcolor{green}3.44 &\cellcolor{green}3.26 &\cellcolor{green}12.19 &\cellcolor{green}5.21 &\cellcolor{green}7.08 &\cellcolor{green}54.81 \\
        \cline{2-12}
        & 0.85 &\cellcolor{green}15.70 &\cellcolor{green}21.46 &\cellcolor{green}10.27 &\cellcolor{green}5.74 &\cellcolor{green}3.37 &\cellcolor{green}3.12 &\cellcolor{green}11.68 &\cellcolor{green}5.18 &\cellcolor{green}7.04 &\cellcolor{green}53.30 \\
        \cline{2-12}
        & 0.80 &\cellcolor{grey}15.70 &\cellcolor{green}21.44 &\cellcolor{green}10.26 &\cellcolor{red}5.75 &\cellcolor{red}3.39 &\cellcolor{green}3.05 &\cellcolor{green}11.33 &\cellcolor{green}5.13 &\cellcolor{green}6.95 &\cellcolor{green}52.16 \\
        \cline{2-12}
        & 0.75 &\cellcolor{green}15.74 &\cellcolor{red}21.48 &\cellcolor{green}10.24 &\cellcolor{green}5.72 &\cellcolor{grey}3.37 &\cellcolor{green}2.92 &\cellcolor{green}10.85 &\cellcolor{green}5.09 &\cellcolor{green}6.89 &\cellcolor{red}54.08 \\

        \specialrule{1.6pt}{1pt}{1pt}
    \end{tabular}
    \caption{The influence of \textbf{Score} on DialoGPT generations for (a) PERSONA-CHAT test set, and (b) ConvAI2 validation set. The green cells indicate that the metric follows the expected trend, i.e., model performance degrades as the score is lowered. The red cells indicate cases where the metric contradicts the expected trend, while the grey cells represent scenarios where there is no change in the value of the metric.}
    \label{tab:scoring}
\end{table*}

\subsubsection{Further Analysis}
\label{subsec:further}

We note that Dist-1 and Dist-2 (the ratio of distinct to total n-grams) for PERSONA-CHAT follow the expected trend and decrease as the score decreases. Since entropy additionally accounts for how evenly n-grams are distributed in the text, we can infer that n-gram distributions become more uniform as the score decreases (while the opposite is desired). We estimate probability distributions for unigrams (Ent-1) and bigrams (Ent-2) present in responses using their relative frequencies. This is done for responses generated using all scores, as well as the target responses in PERSONA-CHAT. Next, we measure how the distributions corresponding to predicted responses (for all scores) differ from the distributions corresponding to target responses. We use the Kullback–Leibler divergence to measure this difference as it is useful for comparing n-gram distributions \citep{kld}. We check for the same trend here, i.e., the value should degrade (higher KL divergence) as scores are lowered. The results shown in Table \ref{tab:ngrams} indicate that the expected trend does occur in most cases, meaning that predicted unigram and bigram distributions differ more from the target distribution as the score decreases. Therefore, we can conclude that greater uniformity in n-gram distributions does not translate to better response capture for PERSONA-CHAT. This can be attributed to the lower diversity of its responses (Ent-1 = 5.68 and Ent-2 = 8.91) compared to ConvAI2 (Ent-1 = 5.79 and Ent-2 = 9.12). Our findings emphasise the need for more diverse dialogue datasets, and improved metrics that take into account reference distributions when measuring diversity in natural language generation.

\begin{table}
    \small
    \centering
    \begin{tabular}{ccc}
         \specialrule{1.6pt}{1pt}{1pt}
         \textbf{Score} & \textbf{Unigrams} & \textbf{Bigrams} \\ 
         \hline
         1.0 & 6.90 & 15.06 \\
         \hline
         0.95 & \cellcolor{green}6.93 & \cellcolor{red}14.97 \\
         \hline
         0.90 & \cellcolor{green}7.01 & \cellcolor{green}15.23 \\
         \hline
         0.85 & \cellcolor{green}7.09 & \cellcolor{green}15.28 \\
         \hline
         0.80 & \cellcolor{green}7.15 & \cellcolor{green}15.30 \\
         \hline
         0.75 & \cellcolor{green}7.25 & \cellcolor{green}15.50 \\
         \specialrule{1.6pt}{1pt}{1pt}
    \end{tabular}
    \caption{KL divergence between probability distributions of unigrams and bigrams in predicted and target responses of the PERSONA-CHAT dataset}
    \label{tab:ngrams}
\end{table}

\section{Discussion}

Our framework revolves around the idea of correlating responses to their quality during training, and then using this knowledge at inference. We present some examples of responses generated using different scores in the supplementary material \citep{supplementary}. While our models do successfully learn the expected trend between responses and their scores (persona-consistency), as evidenced by Tables \ref{tab:scoring} and \ref{tab:ngrams}, this correspondence is not perfect and varies across samples. We also note that most automatic metrics used in this work have been primarily selected due to their widespread adoption for evaluating natural language generation. However, n-gram-based metrics like BLEU fail to capture synonyms and paraphrases in generations \citep{bert-score}. Additionally, diversity metrics are not computed relative to the reference data (Section \ref{subsec:further}) and may therefore not be completely reliable indicators of persona-consistency in dialogue. Future work may explore other metrics that can alleviate these issues.

Furthermore, our data augmentation strategy relies on the idea that nouns capture most persona-relevant information in responses. However, this may not always be true and other parts of speech like verbs, or keywords selected using statistical methods, may also encode persona-relevant information (additional experiments in supplementary material \citep{supplementary}). It is also worth noting that even though POS tagging models have very high accuracy ($\sim$95\% for the English language), they are not perfect, and misclassification may cause persona-relevant nouns to be exempt from masking. We expect this issue to be more pronounced for low-resource languages, which may limit the applicability of our framework.  However, these errors are unlikely to reduce the coherence of augmented samples, since regeneration will likely produce the same part of speech that was masked.

\section{Conclusion}

In this work, we propose an efficient framework for persona-consistent dialogue generation. Our method quantifies response quality using regression-based scores and incorporates these during training. By learning responses conditioned on scores, our dialogue model learns a smooth correlation between responses of varying quality. This is in contrast to only learning golden responses or learning a binary classification between persona-consistent and persona-inconsistent responses. Unlike many previous approaches that perform supervised finetuning and quality alignment separately, we unify both tasks into a single training objective (next token prediction). Experiments with two benchmark datasets show performance boosts for both small and large language models. A natural extension would be to apply our framework to other natural language generation tasks like summarisation or creative writing. Score-based likelihood modification could serve as another avenue for future research.

\section*{Ethics Statement}

This research has used human annotators to assess outputs generated by language models. The protocol for human evaluation has been reviewed and approved by the Engineering and Physical Sciences Faculty Research Ethics Committee, University of Leeds on 14 February 2025 (Reference: EPS FREC - 2024 1910-2422 and EPS FREC - 2025 1910-3236). The participants recruited are graduate students who are native English speakers. They were compensated for their time using the national living wage as the rate, scaled pro rata for the time involved. Before taking part, the evaluators were informed that the data collected would only be used for statistical analysis. All personal data was anonymised for analysis and no identifiable data has been made available. The instructions for participants are given in the supplementary material \citep{supplementary}.



\begin{ack}
AS is supported by a UKRI-funded PhD studentship (Grant Reference: EP/S024336/1). This research made use of the Tier 2 HPC facility JADE2, funded by EPSRC (EP/T022205/1) and the Aire HPC system at the University of Leeds, UK.
\end{ack}



\bibliography{custom}

\end{document}